%% file: main.tex
\documentclass[10pt,twocolumn,letterpaper]{article}

\usepackage{iccv}
\usepackage{times}
\usepackage{epsfig}
\usepackage{graphicx}
\usepackage{amssymb}

\usepackage[utf8]{inputenc} 
\usepackage[T1]{fontenc}    
\usepackage{url}            
\usepackage{booktabs}       
\usepackage{amsfonts}       
\usepackage{nicefrac}       
\usepackage{microtype}      
\usepackage{xcolor}         
\usepackage{algorithm}
\usepackage{algpseudocode}
\usepackage{fp}
\usepackage{wrapfig,lipsum,booktabs,epstopdf}
\usepackage{lipsum}
\usepackage{rotating}
\usepackage{multirow}
\usepackage{algpseudocode}
\usepackage{pifont}
\newcommand{\cmark}{\ding{51}}
\newcommand{\xmark}{\ding{55}}
\usepackage{array, makecell} 
\usepackage{xcolor,colortbl}
\usepackage{comment}
\usepackage{kantlipsum}
\usepackage{tabularx}
\usepackage{booktabs}       
\usepackage{amsmath}
\usepackage{bbm}
\usepackage{diagbox}
\usepackage{amsfonts} 
\usepackage{caption}
\usepackage{etoolbox}
\usepackage{enumitem}
\setlist{topsep=0pt,noitemsep,topsep=0pt,parsep=0pt,partopsep=0pt}
\definecolor{bgreen}{rgb}{0.0, 0.5, 0.0}
\definecolor{deepskyblue}{rgb}{0.0, 0.75, 1.0}

\usepackage[pagebackref=true,breaklinks=true,letterpaper=true,colorlinks,bookmarks=false]{hyperref}

\iccvfinalcopy 

\def\iccvPaperID{9195} 

\ificcvfinal\fi

\begin{document}

\newcommand{\papername}{Image Captioner for Image Captioning Bias Amplification Assessment}
\newcommand{\papernameAbbrev}{ImageCaptioner$^2$}
\newcommand{\benchnameAbbrev}{AnonymousBench}

\title{{\papernameAbbrev}: {\papername}}

\author{
Eslam Mohamed Bakr$^{1}$, 
Pengzhan Sun$^{2}$, 
Li Erran Li$^{3}$, 
Mohamed Elhoseiny$^{1}$ \\
\texttt{\{eslam.abdelrahman, mohamed.elhoseiny\}@kaust.edu.sa} \\
\texttt{pengzhan@comp.nus.edu.sg, lilimam@amazon.com} \\
$^{1}$King Abdullah University of Science and Technology (KAUST) \hspace{0.1cm} \\
$^{2}$National University of Singapore \hspace{0.1cm}
$^{3}$AWS AI, Amazon  \hspace{0.1cm}
}

\maketitle
\ificcvfinal\thispagestyle{empty}\fi

\begin{abstract}
   Most pre-trained learning systems are known to suffer from bias, which typically emerges from the data, the model, or both.
   Measuring and quantifying bias and its sources is a challenging task and has been extensively studied in image captioning. Despite the significant effort in this direction, we observed that existing metrics lack consistency in the inclusion of the visual signal.
   In this paper, we introduce a new bias assessment metric, dubbed \papernameAbbrev, for image captioning.
   Instead of measuring the absolute bias in the model or the data, \papernameAbbrev pay more attention to the bias introduced by the model w.r.t the data bias, termed bias amplification.
   Unlike the existing methods, which only evaluate the image captioning algorithms based on the generated captions only, \papernameAbbrev incorporates the image while measuring the bias.
   In addition, we design a formulation for measuring the bias of generated captions as prompt-based image captioning instead of using language classifiers.
   Finally, we apply our \papernameAbbrev metric across 11 different image captioning architectures on three different datasets, i.e., MS-COCO caption dataset, Artemis V1, and Artemis V2, and on three different protected attributes, i.e., gender, race, and emotions.
   Consequently, we verify the effectiveness of our \papernameAbbrev metric by proposing {\benchnameAbbrev}, which is a novel human evaluation paradigm for bias metrics. 
   Our metric shows significant superiority over the recent bias metric; LIC, in terms of human alignment, where the correlation scores are 80\% and 54\% for our metric and LIC, respectively. 
   The code is available at \href{https://eslambakr.github.io/imagecaptioner2.github.io/}{imagecaptioner2.github.io/}. 
\end{abstract}


\input{sections/1_introduction}

\input{sections/2_related_work}

\input{sections/3_analysis}

\input{sections/4_method}

\input{sections/5_experiment}

\input{sections/6_conclusion}

{\small
\bibliographystyle{ieee_fullname}
\bibliography{egbib}
}

\end{document}

%% file: sections/1_introduction.tex
\section{Introduction}

\begin{figure}
\centering
\captionsetup{font=small}
\includegraphics[width=0.99\linewidth]{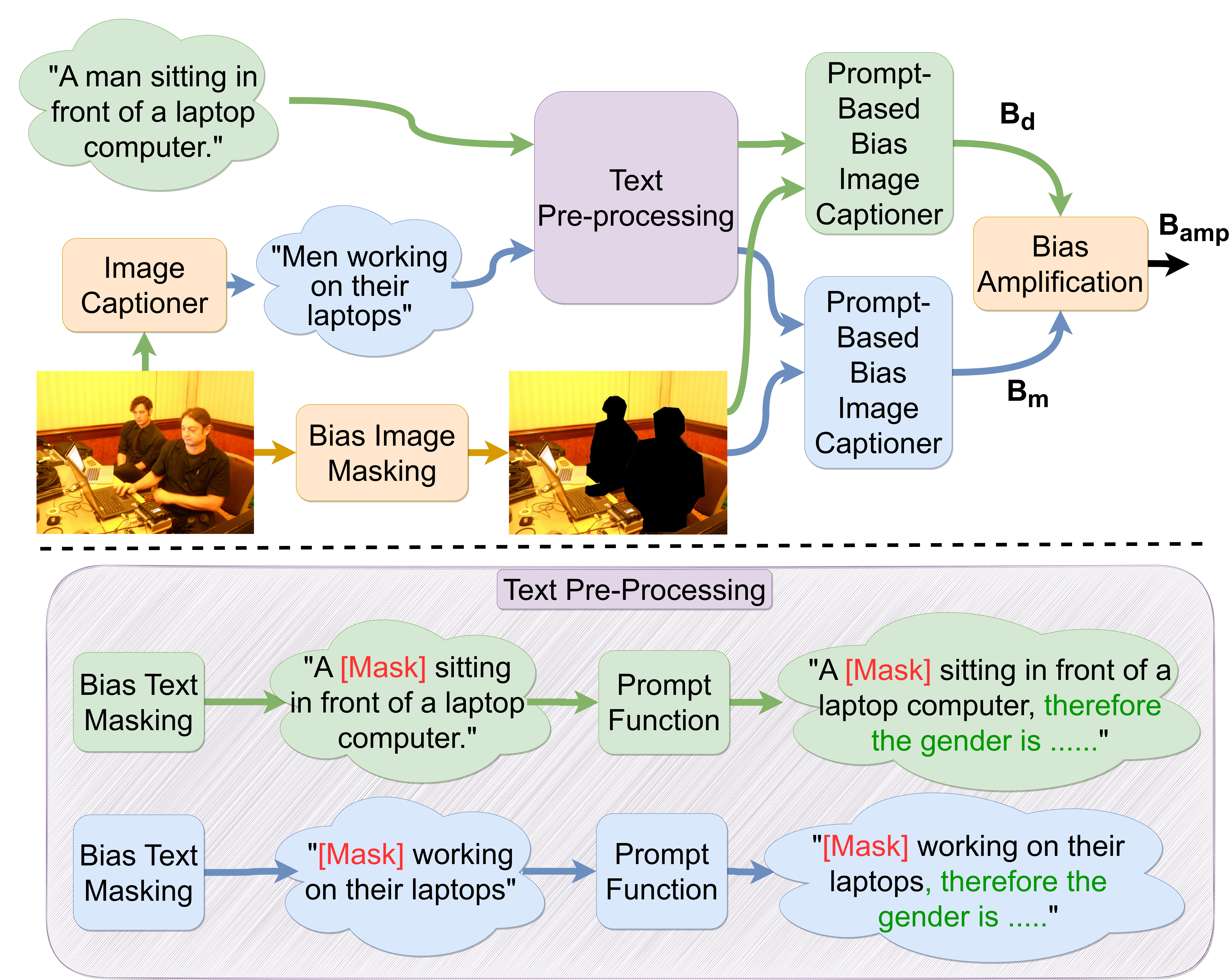}
\caption{An abstract overview for our metric pipeline. 
The GT stream is in \textcolor{green}{green}, and the prediction stream is in \textcolor{blue}{blue}.
Given an input image, the associated GT-caption, and the predicted caption, a text pre-processing module is utilized to refine the captions before feeding them to the prompt-based image captioners. The text pre-processing module performs two main functionalities: 1) Masking the protected-attribute indicative words. 2) Appending the prompt template to the caption. Finally, bias-amplification module is utilized to measure the model bias $B_m$ w.r.t the GT bias $B_d$.}
\label{fig_teaser_fig}
\vspace{-0.4cm}
\end{figure} 

Most deep learning (DL) benchmarks and challenges \cite{deng2009imagenet} \cite{lin2014microsoft} \cite{cirecsan2011high} \cite{kuznetsova2020open} \cite{liao2022kitti} are designed to differentiate and rank different architectures based on accuracy, neglecting other aspects such as fairness.
Nevertheless, measuring the bias and understanding its sources have attracted significant attention recently due to the social impact of the DL models \cite{alvi2018turning} \cite{de2019does} \cite{khan2021one} \cite{stock2018convnets} \cite{thong2021feature} \cite{wang2022revise} \cite{yang2020towards} \cite{du2022vos} \cite{schick2021self}.
For instance, image captioners may learn shortcuts based on correlation, which inevitably suffers from unreliable associations between protected attributes, e.g., gender, and visual or textual clues in the captioning dataset \cite{zhao2017men,wang2019balanced}.

\begin{figure*}
\captionsetup{font=small}
\begin{center}
\includegraphics[width=0.8\linewidth]{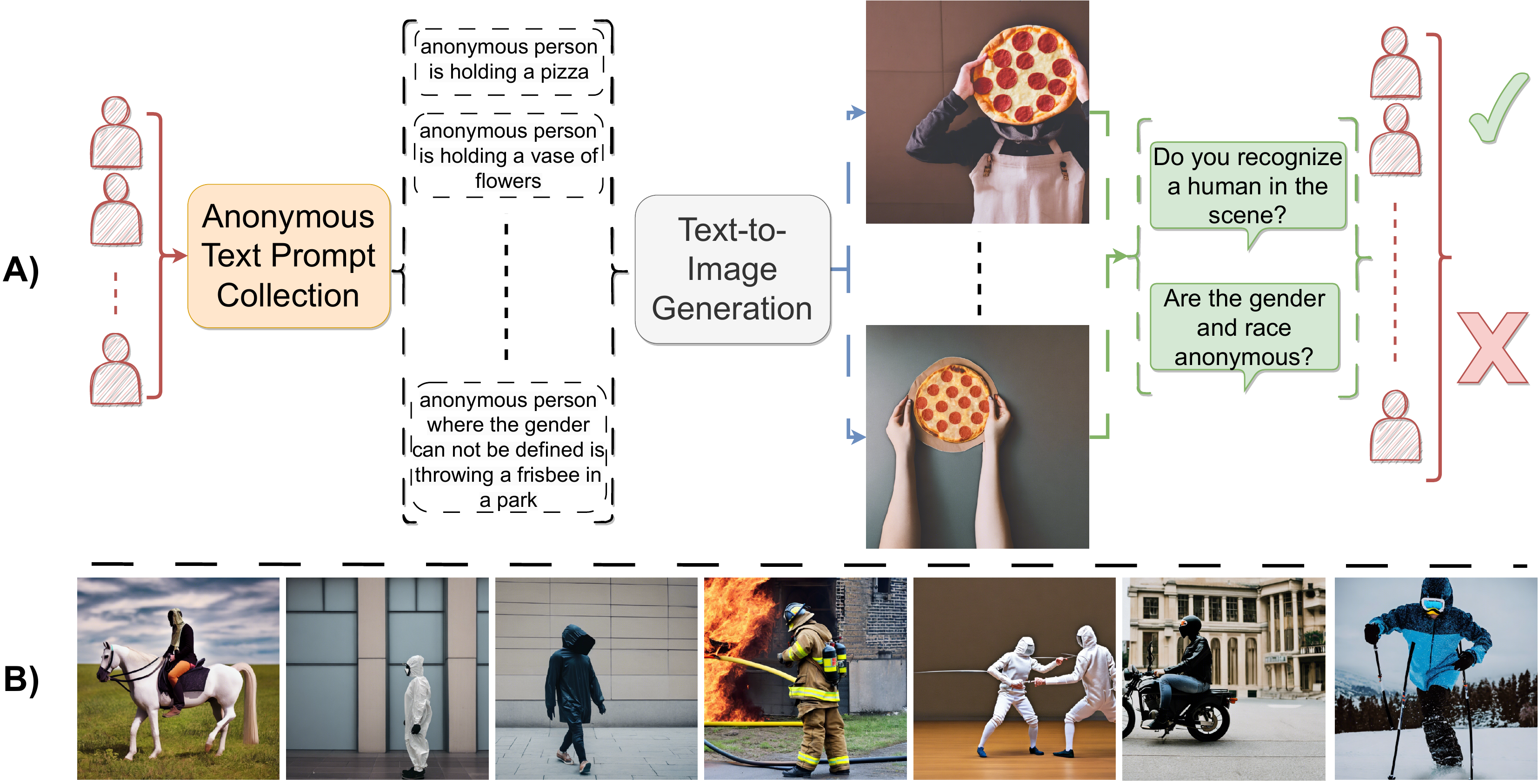}
\end{center}
\caption{{\benchnameAbbrev}. Part A is the detailed pipeline which demonstrates how our novel benchmark is collected. First annotators are asked to write a set of anonymous descriptions, then the generated prompts are fed to a text-to-image model to generate the images. Finally, as a verification step, the same annotators are asked again to filter out the gender or race recognizable images. 
Part B demonstrates some random samples of {\benchnameAbbrev}.}
\label{fig_anonymousbench_pipeline}
\end{figure*}

Recent efforts focus on estimating model bias, driven by the fact that more than balanced data is needed to create unbiased models  \cite{wang2019balanced}.
Bias \cite{zhang2023auditing, bolukbasi2016man, caliskan2017semantics} is characterized by the model's representation of different subgroups when generating the supergroup, such as assessing whether it equally depicts men and women in images of people. 
The primary cause of bias stems from spurious correlations captured during training.
we are interested in the spurious correlation found within ordinary daily scenes, which typically do not evoke bias concerns.
Accordingly, BA \cite{zhao2017men}, DBA \cite{wang2021directional}, and LIC \cite{hirota2022quantifying} measure the model bias w.r.t the bias in the training dataset. In other words, they emphasize the bias introduced by the model regardless of the bias in the ground-truth dataset.
Therefore, detecting and measuring the bias is the first step toward a reliable and accurate image captioner.

BA \cite{zhao2017men} and DBA \cite{wang2021directional}, calculate the co-occurrence frequency of protected bias attribute and detected objects in the picture or some specific words in the caption as a bias score.
An apparent limitation of these evaluations is that only limited visual or language contexts are used for computing the statistics.
Consequently, several learnable bias measurements have been proposed recently \cite{zhao2021understanding,hirota2022quantifying,wang2019balanced}.
The basic assumption of this line of work is that if the bias attribute category, e.g., gender, can be inferred through context even if the attribute feature does not exist, then it indicates that the model is biased.
For instance, the recent work, LIC \cite{hirota2022quantifying}, estimates the bias in image captioning models using trainable language classifiers, e.g., BERT\cite{devlin2018bert}.

Consequently, measuring the bias in image captioning is notably challenging due to the multimodal nature of the task, where the generated text could be biased to the input image and the language model.
Even though significant progress has been achieved in quantifying the bias and its amplification in image captioning models, it remains to be investigated how to properly measure the bias amplified by the image captioning model over the dataset from both the vision and language modalities.

The existing bias metrics formulate the metric as quantifying bias in text generation models, where the image stream is omitted while measuring the bias.
Discarding the visiolinguistic nature of the task in the metric may result in misleading conclusions.
Addressing these limitations, we propose a novel bias metric termed {\papernameAbbrev}, which incorporates the visiolinguistic nature of the image captioning task by utilizing an image captioner instead of a language classifier.

In addition to filling the multi-modalities gap in the existing bias metrics, {\papernameAbbrev} better matches the underlying pre-training model, by proposing a prompt-based image captioner, i.e., the training objective is text generation.
In contrast, the existing bias metric, e.g., LIC, utilizes a language classifier trained for a different objective, i.e., classification objective; this leads to a potential under-utilization of the pre-trained model we assess.
In other words, our proposed metric provides textual prompts that bring the evaluation task closer to the pre-training task.

Experimentally, we evaluate our metric through comprehensive experiments across 11 different image captioning techniques on three different datasets, i.e., MS-COCO caption dataset\cite{lin2014microsoft}, ArtEmis V1 \cite{achlioptas2021artemis}, and ArtEmis V2 \cite{mohamed2022okay}, and on three different protected attributes, i.e., gender, race, and emotions.
In addition, we introduce consistency measures to judge which learnable metric is more consistent against classifiers variations: 1) Introducing conflict-score to count the number of mismatches between different classifiers. 2) Introducing Ranking Consistency, which utilizes Pearson correlation.
In comparison to LIC, we show that based on the consistency measures, our method is more consistent than LIC, where the correlation ratio is 97\% and 92\%, respectively.
Consequently, we verify the effectiveness of our {\papernameAbbrev} metric by proposing {\benchnameAbbrev}, depicted in Figure \ref{fig_anonymousbench_pipeline}, which is a novel human evaluation paradigm for bias metrics. 
Our metric shows significant superiority over the recent bias metric; LIC, in terms of human alignment, where the correlation scores are 80\% and 54\% for our metric and LIC, respectively.

Our contributions can be summarized as follows:

\begin{itemize}

\item We develop a novel bias metric called {\papernameAbbrev}, depicted in Figure \ref{fig_teaser_fig}. To the best of our knowledge, {\papernameAbbrev} is the first metric that is formulated based on the visiolinguistic nature of the image captioning while quantifying the bias.

\item We propose a prompt-based image captioner that better matches the underlying pre-training model we assess.

\item Propose {\benchnameAbbrev} to verify the effectiveness of our {\papernameAbbrev} metric, depicted in Figure \ref{fig_anonymousbench_pipeline}.

\item We propose a new scoring function that matches the objective of measuring the bias.

\item We apply our {\papernameAbbrev} metric across 11 different image captioning techniques on three different datasets, i.e., MS-COCO caption dataset, Artemis V1, and Artemis V2, and on three different attributes, i.e., gender, race, and emotions.
\end{itemize}

%% file: sections/2_related_work.tex
\section{Related Work}

Our contribution draws success from several areas; image captioning,  prompt-based learning, and bias metrics.

\textbf{Image Captioning.}
\cite{kiros2014unifying} proposes the first deep neural language model for solving the image captioning task.
Subsequently, RNN\cite{zaremba2014recurrent, hochreiter1997long} based models, such as NIC\cite{vinyals2015show} and SAT\cite{xu2015show}, together with transformer\cite{devlin2018bert} based models, such as OSCAR\cite{li2020oscar}, are widely accepted because of the impressive performance.
However, image captioners tend to learn shortcuts based on correlation, therefore, researchers have made promising attempts\cite{hendricks2018women,wang2020visual,yang2021causal,yang2021deconfounded,liu2022show} to create unbiased image captioners.
Nevertheless, all these attempts measure the bias from the language stream only.
To fill this gap, we propose a new metric termed \papernameAbbrev, which can capture the bias from both visual and language streams.

\textbf{Prompt-Based Learning.}
Prompt-based methods \cite{liu2021pre, mccann2018natural} reformulate the input $x$ of the downstream task to fit into the training objective of the pre-training phase.
For instance, \cite{kumar2016ask, mccann2018natural, radford2019language, schick2020exploiting} adapt the input $x$ by adding a template $T$ forming a new input $x_p$.
\cite{mccann2018natural} tackles a wide range of NLP tasks in a Question-Answer (QA) fashion. For instance, to tackle the sentiment analysis task, the template $T$ is stacked at the beginning of the input $x$, such as $T = $ "Is this sentence positive or negative", where $x = $ "I watched this movie ten times".
PET \cite{schick2020exploiting} formulate the template as cloze-style phrases to help models understand a given task.

\textbf{Bias Metrics.}
\cite{hendricks2018women} \cite{tang2021mitigating} define a list of protected attributes, e.g., age and gender, and evaluate the accuracy of predicting the specified attributes.
These methods suffer from many drawbacks; 1) They can't handle paraphrasing. 2) Focusing only on the protected attributes and discarding the rest of the caption.
\cite{zhao2021understanding} \cite{wang2019balanced} tries to overcome these problems by incorporating the full caption by utilizing a sentence classifier.
Instead of measuring the absolute bias, BA \cite{zhao2017men}, DBA \cite{wang2021directional}, and LIC \cite{hirota2022quantifying} introduce bias amplification metrics, which measure the relative bias w.r.t the GT datasets.
LIC \cite{hirota2022quantifying} borrows ideas from \cite{wang2019balanced} and \cite{zhao2021understanding}, where it relies on a text classifier to predict the protected attribute.

%% file: sections/3_analysis.tex
\section{Revisiting Image captioning Fairness Metrics }
\label{sec_revist_Fairness_metrics}

In recent years, there have been rapid efforts toward designing robust fairness metrics. 
Nonetheless, all the existing metrics suffer from severe limitations that make them far from being reliable; therefore, in this paper, we take one step toward designing a more robust and reliable metric.
In this section, we first cover the different taxonomies of the existing fairness metrics in image captioning.
Then, we dissect them to show their limitations, which raises several questions, we will address in Section \ref{methodology_section}.

\subsection{Taxonomy of Fairness Metrics}
\label{sec_Fairness_metrics_taxonomy}

We classify the fairness metrics into three categories:

\textbf{Source-Agnostic Vs. Source-Identification metrics.}
Women also Snowboard \cite{hendricks2018women}, and GAIC \cite{tang2021mitigating} assume access to the bias attribute in the predicted captions while measuring the misclassification rate. Accordingly, such metrics can not identify the bias source, which is crucial in designing a debiasing technique.
In contrast, methods that rely on a pre-trained language model to determine whether the model is biased or not, such as \cite{zhao2021understanding}, LIC \cite{hirota2022quantifying}, and \cite{wang2019balanced}, or the methods that rely on calculating the co-occurrences, i.e., BA \cite{zhao2017men} and DBA \cite{wang2021directional}, have the capability of determining the source of the bias.

\textbf{Learnable Vs. Non-Learnable metrics.}
\cite{zhao2021understanding} studies the racial bias, i.e., lighter and darker, by utilizing a dedicated classifier to predict the race based on the predicted caption. In addition, LIC \cite{hirota2022quantifying}, and \cite{wang2019balanced} train additional language classifiers to measure the bias in the data and the model.
In contrast, Women also Snowboard \cite{hendricks2018women}, GAIC \cite{tang2021mitigating}, BA \cite{zhao2017men}, and DBA \cite{wang2021directional} do not utilize any additional learnable parameters to measure the bias.
Women also Snowboard \cite{hendricks2018women} defines the error rate as the number of gender misclassifications, i.e., whether the protected attribute words have been correctly predicted in the generated caption or not.
Consequently, GAIC \cite{tang2021mitigating} formulates the gender bias as the difference in performance between the subgroups of a protected attribute, e.g., GAIC \cite{tang2021mitigating} creates three groups for gender: male, female, and not-specified.
BA \cite{zhao2017men} and DBA \cite{wang2021directional} calculate the co-occurrence between the desired bias attribute, e.g., gender, and the predicted caption.

\textbf{Absolute-Bias Vs. Bias-Amplification.}
Another comparison criterion is whether the metric measures the magnitude of the bias introduced by the model over the bias already existing in the data.
This type of metric can be interpreted as Bias-amplification metrics \cite{zhao2017men} \cite{wang2021directional} \cite{hirota2022quantifying} \cite{wang2019balanced}, where it answers the following question: "Does the model introduce extra bias than ground-truth dataset?"
To this end, these models ground the model bias score to the data bias score. 
In contrast, \cite{zhao2021understanding} \cite{hendricks2018women} \cite{tang2021mitigating} do not provide this valuable information; that's why we call them absolute-bias metrics.

\subsection{Fairness metrics limitations}
\label{sec_Fairness_metrics_limitations}

\tabcolsep=0.08cm
\begin{table*}[t!]
\centering
\caption{Comparison of various bias metrics. Where Full-Context indicates, the whole caption is utilized, not only protected attributes. Multi-Modal reveals the metric incorporates the image alongside the text while measuring the bias of image captioning models. Consistency means the metric gives the same results on multiple runs. Learnable determines whether the metric exploits additional learnable parameters, e.g., language classifiers. Finally, Amplification-Magnitude demonstrates the metric provides information regarding the extra bias introduced by the model over the data.}
\scalebox{0.8}{
\begin{tabular}{cccccccc}
    \toprule[1.5pt]
    \textbf{Method} &  \makecell{\textbf{Women also}\\\textbf{snowbard \cite{hendricks2018women}}}  &  \textbf{GAIC \cite{tang2021mitigating}} & \makecell{\textbf{Understanding} \\\textbf{racial biases \cite{zhao2021understanding}}} &  \textbf{BA \cite{zhao2017men}} & \textbf{DBA \cite{wang2021directional}} & \textbf{LIC \cite{hirota2022quantifying}} & \textbf{Ours}\\
   \midrule[0.75pt]
    \textbf{Full-Context}& \xmark & \xmark & \cmark & \xmark &\xmark & \cmark  & \cmark \\
    \textbf{Multi-Modal}& \xmark & \xmark & \xmark & \xmark &\xmark & \xmark  & \cmark \\
    \textbf{Consistency}&\cmark & \cmark & \cmark & \cmark &\cmark & \xmark  & \cmark \\
    \textbf{Learnable}&\xmark & \xmark & \cmark & \xmark &\xmark & \cmark  & \cmark\\
    \textbf{Amplification Magnitude}&\xmark & \xmark & \xmark & \cmark &\cmark & \cmark  & \cmark\\

  \bottomrule[1.5pt]
\end{tabular}}
\label{tab_bias_metric_comparision}
\end{table*}
Table \ref{tab_bias_metric_comparision} summarizes the existing bias metric limitations.

\textbf{Ignore Multi-Modality.}
Image captioning is a multi-modal task, making evaluating its bias and determining its source more challenging.
However, to the best of our knowledge, non-of the existing image captioning bias metrics \cite{hendricks2018women} \cite{tang2021mitigating} \cite{zhao2021understanding} \cite{zhao2017men} \cite{wang2021directional} \cite{hirota2022quantifying}, as shown in Table \ref{tab_bias_metric_comparision}, include the image while measuring the bias.
All of them degrade the captioning task to a text generation task, which is a flawed approximation.
Driven by this, we propose a novel metric termed \papernameAbbrev, which respects the multiple modalities nature of the image captioning task by including both the image and the text while measuring the bias. In other words, we instead argue for a formulation of measuring the bias of generated caption from the image captioning model as a prompt-based image captioning task.

\textbf{Limited context.}
Women also Snowboard \cite{hendricks2018women}, GAIC \cite{tang2021mitigating}, BA \cite{zhao2017men}, and DBA \cite{wang2021directional} achieve a consistent performance as they use a fixed formula while computing the bias, e.g., co-occurrence, instead of using learnable classifiers, however, they measure the bias based on the appearance of the protected attributes in the caption and discard the rest of the caption in their bias formula.
In contrast, our metric \papernameAbbrev, considers the entire context while measuring the bias and determining its sources.

\textbf{Inconsistency.}
A reliable evaluation metric should have the following characteristics: 1) \emph{Encapsulated}: could be applied to any method without assuming access to its internal parameters or weights.
2) \emph{Consistent and reliable}: must give the same result and conclusion regardless of the multiple runs.
We noticed an inconsistency in LIC \cite{hirota2022quantifying}, when varying the language encoders, where two language encoders are utilized as classifiers; BERT \cite{devlin2018bert}, and LSTM \cite{hochreiter1997long}.
For instance, based on the results reported in the LIC paper \cite{hirota2022quantifying}, when the LSTM classifier is utilized, the LIC score indicates that the Transformer \cite{vaswani2017attention} is better than the UpDn \cite{anderson2018bottom}, where the LIC score is 8.7 and 9, respectively, as the lower LIC score indicates a better model.
While based on the BERT classifier, the UpDn \cite{anderson2018bottom} is much better than the Transformer \cite{vaswani2017attention}, where the LIC score is 4.7 and 5.9, respectively.
We dissect LIC inconsistency in Section \ref{sec_learnable_metrics_consistency}.
This motivates us to propose a more robust metric, termed \papernameAbbrev, which, as shown in Table \ref{tab_bias_metric_comparision}, belongs to the bias-amplification, learnable, and source-identification families.

%% file: sections/4_method.tex
\section{Prompt-Based Bias Amplification Metric}
\label{methodology_section}

Based on the detailed analysis and the limitations discussed in Section \ref{sec_Fairness_metrics_taxonomy} and Section \ref{sec_Fairness_metrics_limitations}, respectively, several questions and concerns arose:
\begin{enumerate}
    \item Can we regard the multi-modality nature by incorporating the image while measuring the bias introduced by the image captioning models? 
    To this end, we utilize the image captioning model to assess the bias introduced by the image captioning model, dubbed \papernameAbbrev, detailed in Section \ref{sec_overview}.
    \item Can we design a bias metric that better matches the underlying pre-training model we assess?
    To this end, we design a formulation of measuring the bias of generated caption as prompt-based image captioning instead of using language classifiers \cite{hirota2022quantifying}, detailed in Section \ref{sec_prompt_based_implementation}.
    \item Can we assess the inner bias of a model without introducing any additional parameters?
    To this end, we show an interesting property of our metric, called Self-Assessment, where we use the same image captioning model we need to assess to measure its own bias, detailed in Section \ref{sec_Self_Assessmen}.
    \item Does the existing scoring function match the objective of measuring the bias? Discussed in Section \ref{sec_Confidence_is_all_you_need}.
\end{enumerate}


\subsection{\papernameAbbrev}
\label{sec_overview}

Driven by the aforementioned limitations, discussed in Section \ref{sec_Fairness_metrics_limitations}, we propose a prompt-based bias amplification metric, termed \papernameAbbrev.

\textbf{Bias-Amplification.}
We not only measure the severity of the bias in predicted captions but also detect the source of the bias. That means, on one hand, we measure the bias from the dataset; on the other hand, we estimate the bias of the caption model.
As shown in Figure \ref{fig_teaser_fig}, two streams are utilized to evaluate an arbitrary image captioning model; 1) ground-truth stream (In green), 2) model stream (In blue).
To measure the caption model bias $B_m$ w.r.t the ground-truth bias $B_d$, a bias-amplification module is utilized, i.e., subtraction operation; $B_{amp}=B_m-B_d$. 
Other relational operator can be utilized, e.g., division, however the division could be thought of as a normalized version of the subtraction; $\frac{B_m}{B_d} \propto \frac{B_m-B_d}{B_d}$.
The normalized version will make it hard to compare performances across different datasets, but provide similar meaning if the dataset is fixed.

\textbf{Text Pre-processing.}
As shown in Figure \ref{fig_teaser_fig}, given an input image, the associated GT-caption, and the predicted caption, a text pre-processing module is utilized to refine the captions before feeding them to the prompt-based image captioners.
The text pre-processing module performs two main functionalities:
1) Masking the protected-attribute indicative words.
2) Appending the prompt template to the caption, detailed in Section \ref{sec_prompt_based_implementation}.
We hypothesize that if the fed captions are not biased, whether GT or predicted captions, then an arbitrary protected-attribute classifier performance should be around the random performance.
Therefore, masking the protected-attribute indicative words is essential to make our hypothesis reasonable.
For instance, if we are concerned by the gender bias, and the input caption is ``A man sitting in front of his laptop computer'', then the output of the masking operation should be ``A [MASK] sitting in front of [MASK] laptop computer'', where all the words that reveal gender information are replaced by mask token.

\textbf{Image Matters.}
As shown in Table \ref{tab_bias_metric_comparision}, all the existing metrics ignore the image stream while measuring the bias.
This motivates us to propose a metric that respects the multi-modality nature by incorporating the input image.
In other words, we design a formulation of measuring the bias in the caption as an image captioning task, not a plain text generation task.
Therefore, in contrast to the existing bias metrics, we are the first work that assesses the image captioning bias using image captioner instead of language classifiers.

\textbf{Bias image masking.}
Consequently, masking the input image is as vital as masking the protected-attribute indicative words.
To ensure our hypothesis is valid, we prevent all leakage sources, i.e., images and text.
We assume access on bounding boxes or segmentation masks to mask the protected attribute visual clues.
However, this assumption is practical as most of the existing image captioning models inherently contain an object detection phase, which we can utilize.
Figure \ref{fig_teaser_fig} demonstrates an example of a masked image.

\subsection{Prompt Implementation}
\label{sec_prompt_based_implementation}

\textbf{Motivation.}
One possible solution to adapt the image captioner to judge whether the model is biased is to stack a classification head.
A key drawback of this solution is that the added classifier is trained for a different objective, i.e., classification objective, rather than the image captioning model objective, i.e., text generation.
In contrast, we introduce a prompt-based metric to better match the underlying pre-training objective of the image captioner that we assess.

\textbf{Prompt engineering.}
To this end, we reformulate the bias measurement by introducing a prompt function $f_{prompt}$, from a classification objective to a text generation objective by refining the input caption by adding a predefined template $T$.
The prompt function $f_{prompt}$ could be implemented in various ways, depending on the position of the empty slot, i.e., [Answer].
The cloze prompt function incorporates the empty slot in the middle of the caption, while the prefix prompt function incorporates it at the end.
We designed our prompt function $f_{prompt}$ in prefix fashion to fit in both language model families, i.e., autoregressive and masked language models (MLM).
For instance, if the protected attribute $a$ is gender, then the template $T$ is incorporated to the masked input $x_m$, such as, $x_p = T? x_m A$, e.g., ``What is the gender of the following sentence? A [MASK] sitting ... computer. [Answer]'', or $x_p = x_m T A$, e.g., ``A [MASK] sitting ... computer. Therefore the gender is [Answer]''.

\textbf{Implementation Details.}
In the training phase, we fine-tune the prompt-based image captioner to predict the last word, i.e., [Answer].
More specifically, given the predicted caption $x_m$, the prompt $T$, and the masked image $I_m$, our prompt-based judge model predicts the answer $A$; the last word, e.g., male or female.
Therefore the training objective is interpreted as follows:
\begin{equation}
    \label{eq_our_training_objective}
    \mathcal{L}^{CE} = log(P(A|x_mT,I_m)).
\end{equation}
Then the bias scores, i.e., $B_m$ and $B_d$, are interpreted as to what extent the model is confident while predicting the answer $A$;
\begin{equation}
    \label{eq_our_bias_score}
    B_{d,m}=\frac{1}{|\mathcal{D}|} \sum_{\left(x_m, I_m\right) \in \mathcal{D}} P(A|x_mT,I_m),
\end{equation}
where $D$ is the GT and predicted captions for $B_d$ and $B_m$, respectively.
During the inference, we deliberately ignore the model predictions by injecting the input refined text as a predicted word at each time-step $t$, which could be interpreted as a teacher-forcing technique \cite{cho2014learning} but with a different intention.
As our main objective is to predict the [Answer] word based on the input caption $x_p$, not the newly predicted caption.

\subsection{Self-Assessment}
\label{sec_Self_Assessmen}

Self-Assessment could be interpreted as a special case of our metric; {\papernameAbbrev}, where the same image captioning model, is used to measure its own bias.
This allows us to fully take advantage of the parameters learned during the pre-training phase without adding any additional computations.
Avoiding adding any new parameters does not only improve efficiency but also avoids adding an extra source of bias, making the evaluation more robust. 

\begin{table*}
\captionsetup{font=scriptsize}
\begin{minipage}[c]{0.26\textwidth}
    \captionof{table}{Ablation study about different scoring functions; Eq. \ref{eq_scoring_function}. Leakage \cite{wang2019balanced} exploits only the indicator function which measures the classification accuracy.
    LIC \cite{hirota2022quantifying} considers also the confidence score alongside the classification accuracy.
    In contrast, we remove the accuracy measure, and define the bias score as only the confidence score, as shown in Eq. \ref{eq_scoring_function}.}
    \label{tab_ablation_scoringfunction}
    \scalebox{0.7}{
    \begin{tabular}{ c c c c }
        \toprule[1.5pt]
        Method & Leakage $\downarrow$ & LIC $\downarrow$ & Ours $\downarrow$ \\
        \midrule[0.75pt]
        NIC \cite{vinyals2015show} & -0.47 & 1.84 & 2.68 \\
        SAT \cite{xu2015show} & -0.47 & 1.96 & 1.64 \\
        FC \cite{rennie2017self} & -0.61 & 1.56 & 6.29 \\
        Att2in \cite{rennie2017self} & -0.51 & 3.07 & 6.17 \\
        UpDn \cite{anderson2018bottom} & -0.65 & 1.24 & 6.64 \\
        Trans. \cite{vaswani2017attention} & 1.06 & 1.39 & 6.19 \\
        Oscar \cite{li2020oscar} & -0.56 & 1.58 & 5.09 \\
        NIC+ \cite{hendricks2018women} & -0.47 & 2.91 & 3.18 \\
        NIC+Equ. \cite{hendricks2018women} & -0.56 & 0.33 & 5.93 \\
        \bottomrule[1.5pt]
    \end{tabular}
    }
\end{minipage}
\hfill
\begin{minipage}[c]{0.40\textwidth}
    \centering
    \captionof{table}{The bias amplification results for the gender attribute on MS-COCO dataset.
    The ratio and the error are introduced in \cite{hendricks2018women}.
    The ratio is definded based on the number of sentences which belong to a female set to sentences which belong to a male set.
    The error rate is the number of gender misclassifications.
    BA \cite{zhao2017men} and DBA \cite{wang2021directional} measure the bias based on the appearance of the protected attributes in the caption, i.e., co-occurrence.}
    \label{tab_benchmark_gender_coco}
    \scalebox{0.65}{
    \begin{tabular}{ c c c c c c c c }
    \toprule[1.5pt]
        Method & LIC $\downarrow$ & Ratio $\downarrow$ & Error $\downarrow$ & BA $\downarrow$ & $DBA_{G} \downarrow$ & $DBA_{O} \downarrow$ & Ours $\downarrow$\\
        \midrule[0.75pt]
        NIC \cite{vinyals2015show} & 3.7 & 2.47 & 14.3 & 4.25 & 3.05 & 0.09 & 2.68 \\
        SAT \cite{xu2015show} & 5.1 & 2.06 & 7.3 & 1.14 & 3.53 & 0.15 & 1.64 \\
        FC \cite{rennie2017self} & 8.6 & 2.07 & 10.1 & 4.01 & 3.85 & 0.28 & 6.29 \\
        Att2in \cite{rennie2017self} & 7.6 & 2.06 & 4.1 & 0.32 & 3.60 & 0.29 & 6.17 \\
        UpDn \cite{anderson2018bottom} & 9.0 & 2.15 & 3.7 & 2.78 & 3.61 & 0.28 & 6.64 \\
        Trans. \cite{vaswani2017attention} & 8.7 & 2.18 & 3.6 & 1.22 & 3.25 & 0.12 & 6.19 \\
        OSCAR \cite{li2020oscar} & 9.2 & 2.06 & 1.4 & 1.52 & 3.18 & 0.19 & 5.09 \\
        NIC+ \cite{hendricks2018women} & 7.2 & 2.89 & 12.9 & 6.07 & 2.08 & 0.17 & 3.18 \\
        NIC+Equ. \cite{hendricks2018women} & 11.8 & 1.91 & 7.7 & 5.08 & 3.05 & 0.20 & 5.93 \\
        \bottomrule[1.5pt]
    \end{tabular}
    }
\end{minipage}
\hfill
\begin{minipage}[c]{0.32\textwidth}
    \captionof{table}{The bias amplification results for the gender attribute on MS-COCO datasets for LIC and our metric using different judge models.
    The down arrows indicates less is better.
    The ranking of captioning models is reported in \textcolor{red}{red}, which indicates to what extend the metric is consistent when changing the judging model.}
    \label{tab_inconsistency_comparison}
    \scalebox{0.7}{
    \begin{tabular}{ c c c c c }
    \toprule[1.5pt]
        \multirow{2}{*}{Method} & \multicolumn{2}{c}{LIC $\downarrow$} & \multicolumn{2}{c}{Ours $\downarrow$} \\ 
						& LSTM & BERT & SAT & GRIT  \\ 
        
        \midrule[0.75pt]
        NIC \cite{vinyals2015show}& 3.7 \textcolor{red}{(1)} & -0.8 \textcolor{red}{(1)} & 2.68 \textcolor{red}{(2)} & 1.02 \textcolor{red}{(2)}\\
        SAT \cite{xu2015show} & 5.1 \textcolor{red}{(2)} & 0.3 \textcolor{red}{(2)} & 1.64 \textcolor{red}{(1)} & 0.61 \textcolor{red}{(1)} \\
        FC \cite{rennie2017self} & 8.6 \textcolor{red}{(5)} & 2.9 \textcolor{red}{(5)} & 6.29 \textcolor{red}{(8)} & 2.30 \textcolor{red}{(5)} \\
        Att2in \cite{rennie2017self} & 7.6 \textcolor{red}{(4)} & 1.1 \textcolor{red}{(3)} & 6.17 \textcolor{red}{(6)} & 2.78 \textcolor{red}{(6)}\\
        UpDn \cite{anderson2018bottom} & 9.0 \textcolor{red}{(7)} & 4.7 \textcolor{red}{(6)} & 6.64 \textcolor{red}{(9)} & 2.82 \textcolor{red}{(7)}\\
        Trans. \cite{vaswani2017attention} & 8.7 \textcolor{red}{(6)} & 5.9 \textcolor{red}{(8)} & 6.19 \textcolor{red}{(7)} & 2.90 \textcolor{red}{(8)}\\
        OSCAR \cite{li2020oscar} & 9.2 \textcolor{red}{(8)} & 4.9 \textcolor{red}{(7)} & 5.09 \textcolor{red}{(4)} & 2.21 \textcolor{red}{(4)}\\
        NIC+ \cite{hendricks2018women} & 7.2 \textcolor{red}{(3)} & 1.8 \textcolor{red}{(4)} & 3.18 \textcolor{red}{(3)} & 1.16 \textcolor{red}{(3)}\\
        NIC+Equ. \cite{hendricks2018women} & 11.8 \textcolor{red}{(9)} & 7.3 \textcolor{red}{(9)} & 5.93 \textcolor{red}{(5)} & 3.08 \textcolor{red}{(9)}\\
        \bottomrule[1.5pt]
    \end{tabular}
    }
\end{minipage}
\end{table*}

\subsection{Confidence is all you need}
\label{sec_Confidence_is_all_you_need}

The leakage \cite{wang2019balanced} and LIC \cite{hirota2022quantifying} estimate the gender bias for image classification and image captioning, respectively, using external classifiers $f_{cls}$, depicted in Eq. \ref{eq_bias_scoring_function}.

\begin{equation}
    \mathbf{B_d} =\frac{1}{|\mathcal{D}|} \sum_{\left(y, a\right) \in \mathcal{D}} f_s(y, a).
    \label{eq_bias_scoring_function}
\end{equation}

Consider a sample $(I, y, a)$, where $I$ is the input image, $y$ is the corresponding output, e.g., caption in image caption task, and $a$ is the protected attribute, e.g., gender, both of them \cite{wang2019balanced} \cite{hirota2022quantifying} use a scoring function $f_s$ that mixes up the accuracy objective with the bias objective.
As shown in Eq. \ref{eq_scoring_function}, the leakage \cite{wang2019balanced} exploits only the indicator function which measures the classification accuracy.
Consequently, LIC \cite{hirota2022quantifying} argues that the uncertainty measure provides additional evidence for measuring bias, so LIC considers also the confidence score, as shown in Eq. \ref{eq_scoring_function}.

\begin{equation}
    \label{eq_scoring_function}
    f_s(y, a) =
    \begin{cases}
    \mathbbm{1}[f_{cls}(y)=a] & Leakage\\
    S_a(y)* \mathbbm{1}[f_{cls}(y)=a] & LIC\\
    S_a(y) & Ours
    \end{cases}.
\end{equation}

In contrast, aligned with the bias objective, we remove the accuracy measure represented in the indicator function, and define the bias score as only the confidence score, i.e., $B_{d,m} = S_a(y)$ following Eq. \ref{eq_our_bias_score}.
The assumption of such adjustment can be shown in the following case, taking the gender bias measurement as an example, where the protected attribute $a=female$.
If the output confidence score $S_a(y)$ for male and female are 0.51 and 0.49, this means that the attribute classifier $f_{cls}$ predicts \textit{male}.
According to the indicator function this given sample will be discarded, even if we know that the confidence score near 0.5 means there is no bias.
Driven by this, we argue that confidence is all you need for bias scoring.

%% file: sections/5_experiment.tex
\section{Experiments}

\subsection{Datasets and Models}

We evaluate our proposed metric, {\papernameAbbrev}, on two datasets, i.e., MS-COCO captioning dataset \cite{lin2014microsoft} for gender and race attributes, Artemis V1 \cite{achlioptas2021artemis}, and V2 \cite{mohamed2022okay} for emotions attribute. 
For gender and race, we validate the effectiveness of our metric on a wide range of image captioning models, i.e., NIC \cite{vinyals2015show}, SAT \cite{xu2015show}, FC \cite{rennie2017self}, Att2in \cite{rennie2017self}, UpDn \cite{anderson2018bottom}, Transformer \cite{vaswani2017attention}, OSCAR \cite{li2020oscar}, NIC+ \cite{hendricks2018women}, and NIC+Eq \cite{hendricks2018women}.
While for Artemis V1 \cite{achlioptas2021artemis} and V2 \cite{mohamed2022okay}, we explor SAT \cite{xu2015show}, and Emotion-Grounded SAT (EG-SAT) \cite{achlioptas2021artemis} with its variants.
The EG-SAT is an adapted version of SAT that incorporates the emotional signal into the speaker to generate controlled text.
Two variants of EG-SAT are studied based on the source of the emotion:
1) Img2Emo. A pre-trained image-to-emotion classifier is utilized to predict the emotion.
2) Voting. Each image has, on average, eight different captions; therefore, the input emotion is conducted by a simple voting mechanism to pick the most frequent emotion in the GT captions. 

\subsection{Implementation Details}

\textbf{Masking Attribute Clues.}
To measure gender bias, we mask attribute-related information in both image and text.
Specifically, for gender and racial bias, the image is masked using GT masks or predicted bboxes.
In addition, the protected attribute-related words are replaced by a [MASK] token.
In contrast, we do not mask any image region when focusing on emotion bias, as emotion attributes are not represented directly in artworks. Instead, we only mask sentimental words that leak the emotion attribute in captions.

\textbf{Network and Training Configuration.}
We explore two different prompt-based image captioning models, i.e., SAT \cite{xu2015show} and GRIT \cite{nguyen2022grit} to act as classifiers.
To train our prompt-based image captioner, we split the original validation set, around $10^4$ images, into training, validation, and testing sets, i.e., 70\%, 10\%, and 20\%, respectively.
These splits are balanced based on the protected attribute.
All captioning models are trained using the same training configurations mentioned in \cite{hirota2022quantifying}.
The added template $T$ is ``Therefore, the gender is [Answer]'', "Therefore, The race is [Answer]" and ``Therefore, the emotion is [Answer]'' for gender, race, and emotions, respectively.
We train our prompt-based image captioner for 40 epochs from scratch, using the weight initialization strategy described in \cite{he2015delving}.
Adam optimizer \cite{kingma2014adam} and mini-batch size of 128 are used for training all our models.

\textbf{Software and Hardware Details.}
Our metric is implemented in Python using the PyTorch framework.
All experiments are conducted using four NVIDIA V100 GPUs.

\subsection{Scoring Function Ablation Study}

As shown in Table \ref{tab_ablation_scoringfunction}, the scoring function heavily influences the results.
As discussed in Section \ref{sec_Confidence_is_all_you_need}, the uncertainty or confidence score of the model is a direct reflection of the severity of bias. 
The higher confidence score of the model predictions indicates more severe bias.
Leakage, LIC, and \papernameAbbrev \ share the same model weights, however, it is hard to get reasonable and insightful observations from the result of Leakage as most models achieve the same bias score; almost -0.5.
Consequently, when the confidence score is added as in LIC, the bias scores influenced a lot, which indicates that confidence has the dominant role, which is consistent with our arguments in Section \ref{sec_Confidence_is_all_you_need}.
This supports the importance of removing the accuracy indicator from the scoring function.
Therefore, we adapt the scoring function to rely only on the confidence score, Eq. \ref{eq_bias_scoring_function}.

\subsection{Gender Bias Benchmark on MS-COCO}

We benchmark our metric, \papernameAbbrev, in terms of the gender bias across a wide range of captioning models, against the existing bias metrics.
For fair comparison, we follow the training configurations mentioned in \cite{hirota2022quantifying}.
Based on our metric, all captioning models amplify the bias, which is consistent with other bias metrics.
In addition, we observe the same observation presented in \cite{hirota2022quantifying} and \cite{wang2021directional}, where the NIC-Equ amplifies the gender bias despite enhancing the classification accuracy.
Interestingly, the equalizer almost tripled the bias score compared to NIC based on LIC. In contrast, the bias score is only doubled based on our measure.
This decrease in the gap between NIC and NIC-Equ between LIC and our metric, 3x and 2x, respectively, is reasonable as the Equalizer allows models to make accurate predictions of the gender based on the visual region of the person.
Therefore, intuitively, the equalizer effect will be included in our metric as it respects the multi-modality nature of the task, as discussed in Section \ref{sec_overview}.

\subsection{Which metric is better?}

\begin{figure}
\captionsetup{font=scriptsize}
\begin{minipage}[c]{0.2\textwidth}
\begin{center}
\includegraphics[width=0.95\textwidth]{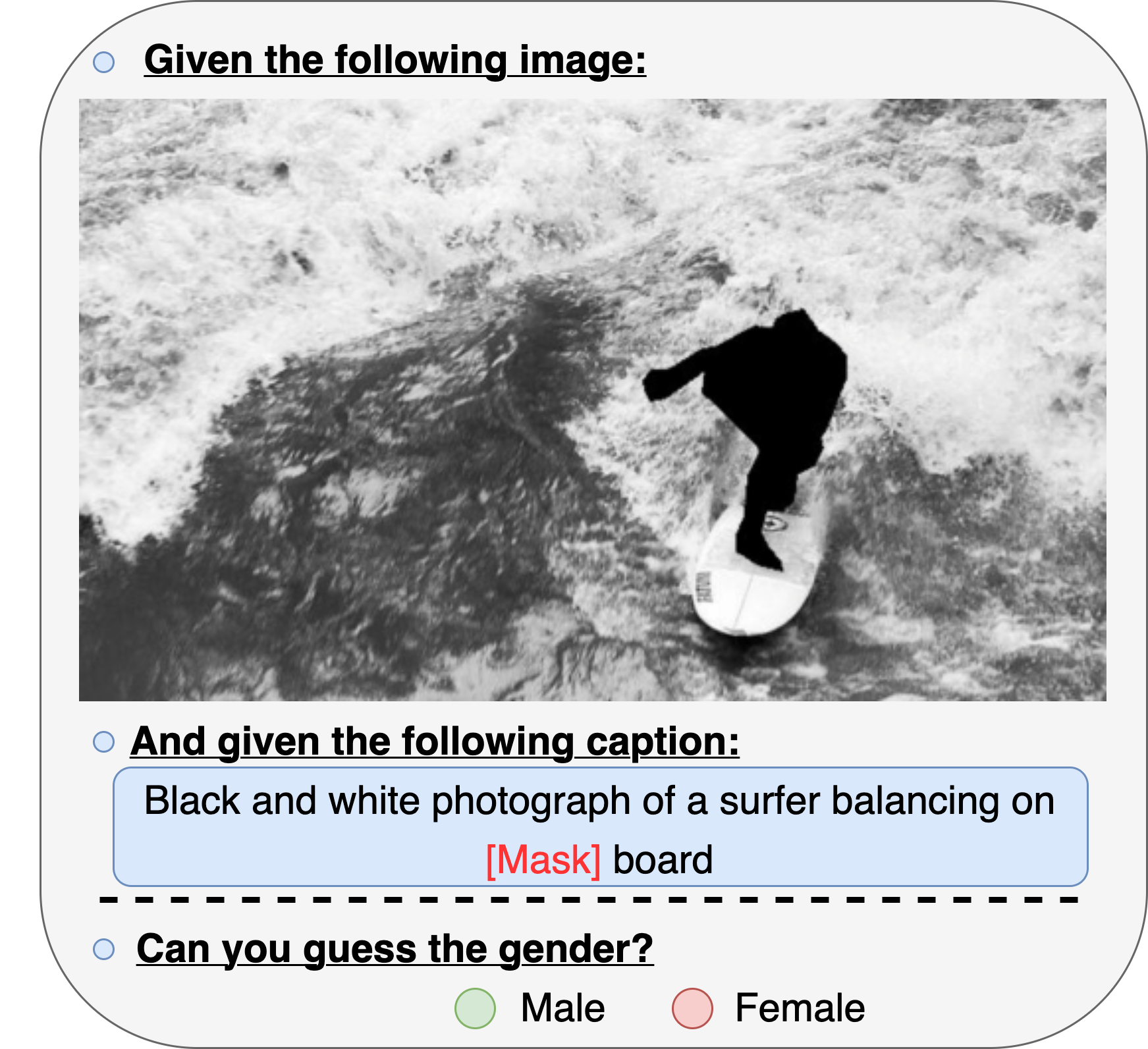}
\captionof{figure}{Human evaluation UI. The annotators try to guess the gender of the masked person.}
\label{fig_amt_ui}
\end{center}
\end{minipage}
\hfill
\hfil
\begin{minipage}[c]{0.27\textwidth}
\begin{center}
\includegraphics[width=1.0\textwidth]{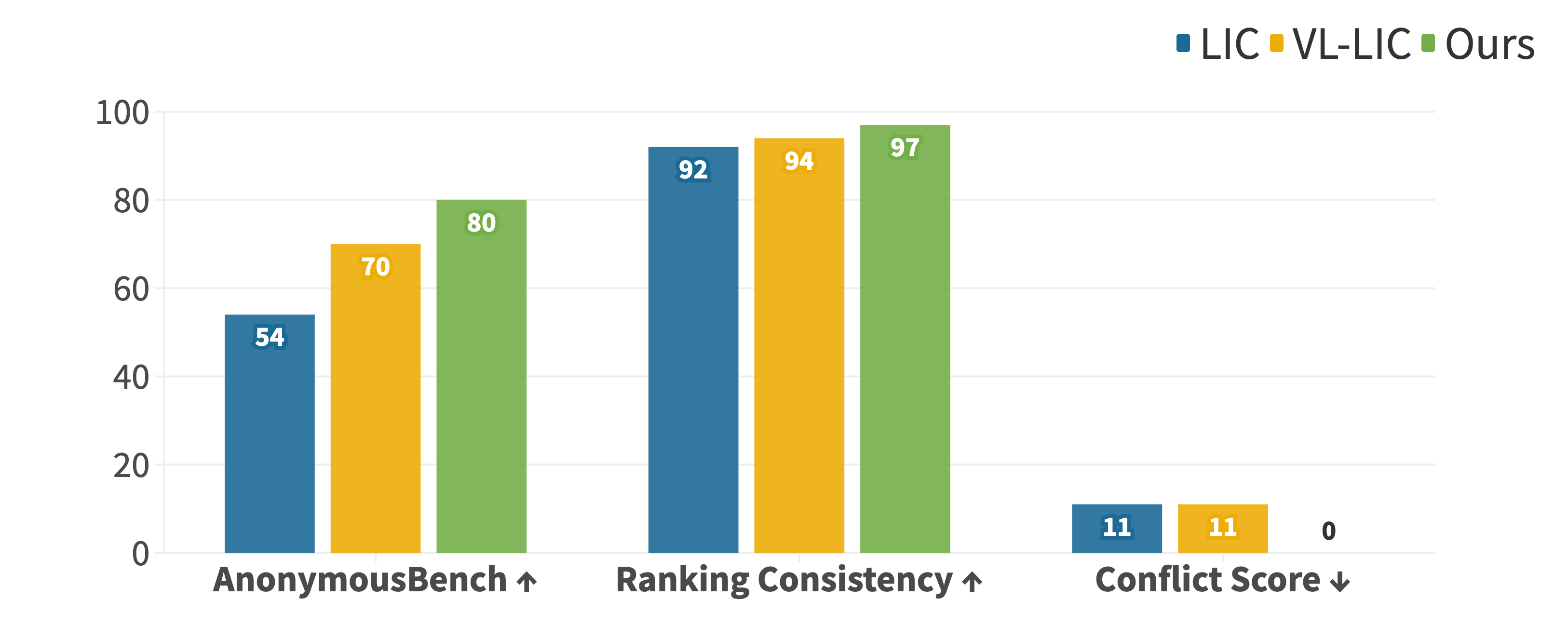}
\captionof{figure}{Comparison between our metric, LIC, and the multi-modal variant of LIC, termed VL-LIC, in terms of human evaluation using our AnonymousBench, ranking consistency, and conflict score.}
\label{fig_human_corr_conflict}
\end{center}
\end{minipage}
\vspace{-0.3cm}
\end{figure}

Despite the superiority of our method and LIC intuitively over the existing metrics as depicted in Section \ref{sec_Fairness_metrics_limitations} and Section \ref{sec_overview}, the ranking inconsistency among different bias metrics raises the importance of developing a method to evaluate the effectiveness of each bias metric.
Consequently, this raises the importance of developing a method to evaluate the effectiveness of each bias metric quantitatively.
Especially our metric and LIC. As both of them are learnable metrics, measure the amplification magnitude, and include the full-context while measuring the bias.

\tabcolsep=0.18cm
\begin{table*}
\captionsetup{font=scriptsize}
\begin{minipage}[c]{0.22\textwidth}
    \centering
    \captionof{table}{Human evaluation results on {\benchnameAbbrev}.}
    \label{tab_anonymousbench}
    \scalebox{0.6}{
    \begin{tabular}{cccc}
        \toprule
        Method & LIC $\downarrow$ & Ours $\downarrow$ & GT $\downarrow$\\
        \midrule
        NIC  \cite{vinyals2015show}          & 97.5     & 95.8 & 74.0 \\
        SAT  \cite{xu2015show}               & 92.0     & 94.0 & 33.0 \\
        FC  \cite{rennie2017self}            & 94.5     & 94.8 & 48.5 \\
        Att2in \cite{rennie2017self}         & 93.0     & 92.3 & 48.5 \\
        UpDn \cite{anderson2018bottom}       & 96.0     & 95.2 & 78.5 \\
        Trans. \cite{vaswani2017attention}   & 99.0     & 95.0 & 63.0 \\
        OSCAR \cite{li2020oscar}             & 96.5     & 91.5 & 27.0 \\
        NIC+ \cite{hendricks2018women}       & 97.5     & 95.6 & 87.5 \\
        NIC+Eq \cite{hendricks2018women}     & 94.5     & 94.7 & 61.5 \\
        \bottomrule
    \end{tabular}
    }
\end{minipage}
\hfil
\begin{minipage}[c]{0.25\textwidth}
    \centering
    \captionof{table}{Comparison of various detectors against the GT masks using {\papernameAbbrev}.}
    \label{tab_detection_bias}
    \scalebox{0.6}{
    \begin{tabular}{cccc}
        \toprule
        Method & GT-Masks  & YOLOX & MaskRCNN \\
        \midrule
        NIC \cite{vinyals2015show}           & 2.7     & 2.9   & 2.7 \\
        SAT \cite{xu2015show}                & 1.6     & 1.4   & 1.8 \\
        FC \cite{rennie2017self}             & 6.3     & 6.5   & 6.5 \\
        Att2in \cite{rennie2017self}         & 6.2     & 6.3   & 6.2 \\
        UpDn \cite{anderson2018bottom}       & 6.6     & 7.1   & 6.9 \\
        Trans. \cite{vaswani2017attention}   & 6.2     & 6.1   & 6.4 \\
        OSCAR \cite{li2020oscar}             & 5.1     & 5.4   & 5.2 \\
        NIC+  \cite{hendricks2018women}      & 3.2     & 3.3   & 3.6 \\
        NIC+Eq  \cite{hendricks2018women}    & 5.9     & 6.0   & 6.2 \\
        \bottomrule
    \end{tabular}
    }
\end{minipage}
\hfil
\hfil
\begin{minipage}[c]{0.47\textwidth}
    \centering
    \captionof{table}{Emotion bias on Artemis datasets \cite{achlioptas2021artemis} \cite{mohamed2022okay}, where EG-SAT stands for emotional-grounding SAT. Concretely for the emotion input of EG-SAT, Img2Emo indicates the grounding emotion is predicted by a pre-trained image-to-emotion classifier; Voting means the grounding emotion is selected by a majority vote from multiple GT emotion labels of the artwork sample.}
    \label{tab_emotion_benchmark}
    \scalebox{0.6}{
    \begin{tabular}{ c c c c c c c c c}
        \toprule[1.5pt]
        \makecell{\textbf{Artemis}\\\textbf{Version}} & \textbf{Method} & \makecell{\textbf{Emotion}\\\textbf{Source}} & \multicolumn{3}{c}{\textbf{LIC} $\downarrow$} &  \multicolumn{3}{c}{\textbf{Ours} $\downarrow$} \\
        \cmidrule(r){4-6}
        \cmidrule(r){7-9}
        \- & \- & \- & \textbf{M} & \textbf{D} & \textbf{Amp.} & \textbf{M} & \textbf{D} & \textbf{Amp.}\\
        \cmidrule(r){1-3}
        \cmidrule(r){4-6}
        \cmidrule(r){7-9}
        \multirow{3}{*}{V1} & SAT & N/A & 15.01 & 13.85 & \textbf{1.16} & 87.23 & 85.04 & \textbf{2.19}\\
        \- & EG-SAT & Img2Emo & 60.38 & 13.85 & \textbf{46.53} & 95.43 & 85.04 & \textbf{10.39}\\
        \- & EG-SAT & Voting & 45.08 & 13.85 & \textbf{31.23} & 88.66 & 85.04 & \textbf{3.62}\\
        \midrule[0.75pt]
        \multirow{3}{*}{V2} & SAT & N/A & 37.08 & 15.34 & \textbf{21.74} & 88.70 & 80.00 & \textbf{8.7}\\
        \- & EG-SAT & Img2Emo & 68.17 & 15.34 & \textbf{52.83} & 93.24 & 80.00 & \textbf{13.24}\\
        \- & EG-SAT & Voting & 50.37 & 15.34 & \textbf{35.03} & 91.18 & 80.00 & \textbf{11.18}\\
        \bottomrule[1.5pt]
    \end{tabular}
    }
\end{minipage}
\end{table*}

\subsubsection{Human Evaluation}
\label{sec_human_eval}

To fairly compare the different bias evaluation metrics, a human evaluation has to be conducted. However, it is hard design such an evaluation for the bias. For instance, Figure \ref{fig_amt_ui} demonstrates one possible solution, which is asking the annotators to try to guess the gender given the AI generated captions. Unfortunately, the formulation will be not accurate enough as humans already biased, therefore this approach could lead to human bias measurement instead of metric evaluation. To tackle this critical point, we introduce {\benchnameAbbrev}, depicted in Figure \ref{fig_anonymousbench_pipeline}.

To prove the effectiveness of our method, we propose {\benchnameAbbrev}, Figure \ref{fig_anonymousbench_pipeline}; gender and race agnostic benchmark that consists of 1k anonymous images. First, we ask annotators to write 500 text prompts about various scenes with anonymous people. Secondly, Stable-Diffusion V2.1 \cite{rombach2022high} is utilized to generate ten images per prompt resulting in 5k images.
To address any potential bias \cite{zhang2023auditing, bakr2023hrs} in the generated images, a human evaluation was conducted to filter out the non-agnostic images based on two simple questions; 1) Do you recognize a human in the scene? 2) If yes, Are the gender and race anonymous? Finally, the filtered images, 1k images, are fed to each model, which we assess, to generate the corresponding captions. We expect that the best model would instead predict gender-neutral words, e.g., person instead of man or woman, as the gender is not apparent. Therefore, the GT score is defined based on whether a human can guess the gender from the generated captions and averaged across the whole data. 
Table \ref{tab_anonymousbench} demonstrates LIC, {\papernameAbbrev}, and GT results on our proposed benchmark; {\benchnameAbbrev}. 
The Pearson correlation is employed to measure the alignment between the metrics and human evaluation. 
As shown in Figure \ref{fig_human_corr_conflict}, our metric is more aligned with the human evaluation, where the correlation scores are 80\% and 54\% for our metric and LIC, respectively.

\subsubsection{Learnable Metrics Consistency}
\label{sec_learnable_metrics_consistency}

As discussed in Section \ref{sec_Fairness_metrics_taxonomy}, learnable metrics utilize additional language classifiers to measure the bias.
Consequently, it may be inconsistent across different classifiers.
To measure the consistency, LIC \cite{hirota2022quantifying} relies on the agreement between different classifiers on the best and the worst models in terms of bias.
Following such a naive approach may lead to an inadequate conclusion.
For instance, we observe an inconsistency in ranking, i.e., shown in red parenthesis in Table \ref{tab_inconsistency_comparison}.
In addition, there is no agreement on whether the model is bias-amplified, e.g., NIC based on LSTM amplifies the bias (3.7) opposed to BERT (-0.8), where a positive score indicates the model amplifies the bias.

Consequently, this motivates us to introduce two consistency measures to judge which learnable metric is classifier invariant; more consistent against classifier variations. 
1) The conflict score (CS): Count the number of conflicts when changing the classifier.
2) Ranking consistency (RC): Measure the correlation between different classifiers.



\textbf{Conflict Score.}
A reliable metric should be classifier invariant, at least in terms of the conclusion, i.e., whether the model is biased.
Therefore, we introduce a conflict score to measure the percentage of mismatches between different classifiers, where the less conflict score is better.
First, we classify the models into binary categories, biased and not biased, where positive means the model is biased.
Then, we calculate the miss-classification rate among different classifiers.
Based on results reported in Table \ref{tab_inconsistency_comparison} and Figure \ref{fig_human_corr_conflict}, our metric, \papernameAbbrev, is more robust, as the conflict score is 0 and 11.11 \% for ours and LIC, respectively.

\textbf{Ranking Consistency.}
To measure the ranking consistency, we utilize the Pearson correlation.
Aligned with the conflict score, the correlation score, depicted in Figure \ref{fig_human_corr_conflict}, also probes that our metric is more robust than LIC, where the correlation score based on the results demonstrated in Table \ref{tab_inconsistency_comparison} is 97\% and 92\% for ours and LIC, respectively.
Where a higher correlation score indicates a more consistent metric.

\subsection{Replacing GT Masks by Detectors}

We replace GT masks, using two off-the-shelf detectors to predict the humans to mask them from the image. As shown in Table \ref{tab_detection_bias}, both YOLOX \cite{ge2021yolox} and MaskR-CNN \cite{he2017mask} almost achieve the same bias score as the GT segmentation masks. In addition, they are fully correlated, which indicates that the detection phase does not introduce additional bias.

\subsection{Multi-Modal Classifier}

To highlight the impact of incorporating the image while measuring the bias, we implement a simple vision-and-language classifier, termed VL-LIC, as a baseline.
Accordingly, we have integrated ResNet50 as the visual backbone in conjunction with the BERT language encoder utilized by LIC.
As depicted in Figure \ref{fig_human_corr_conflict}, incorporating the image, referred to as VL-LIC, demonstrates an enhancement in human correlation on our AnonymousBench and improved consistency compared to LIC, which relies solely on language.
Furthermore, including the image lies in accessing a richer signal and capturing spurious correlations more accurately, as the generated caption alone may not provide comprehensive details encompassed by the image.

However, the prompt-based metric remains superior due to its alignment with the underlying pre-trained model, which follows an auto-regressive approach by predicting subsequent words based on preceding ones.
Furthermore, employing pre-trained vision and language classifiers as judge models introduces additional bias, thus posing challenges in drawing robust conclusions.

\subsection{Qualitative results}

\begin{figure}
\captionsetup{font=small}
\begin{center}
\includegraphics[width=0.8\linewidth]{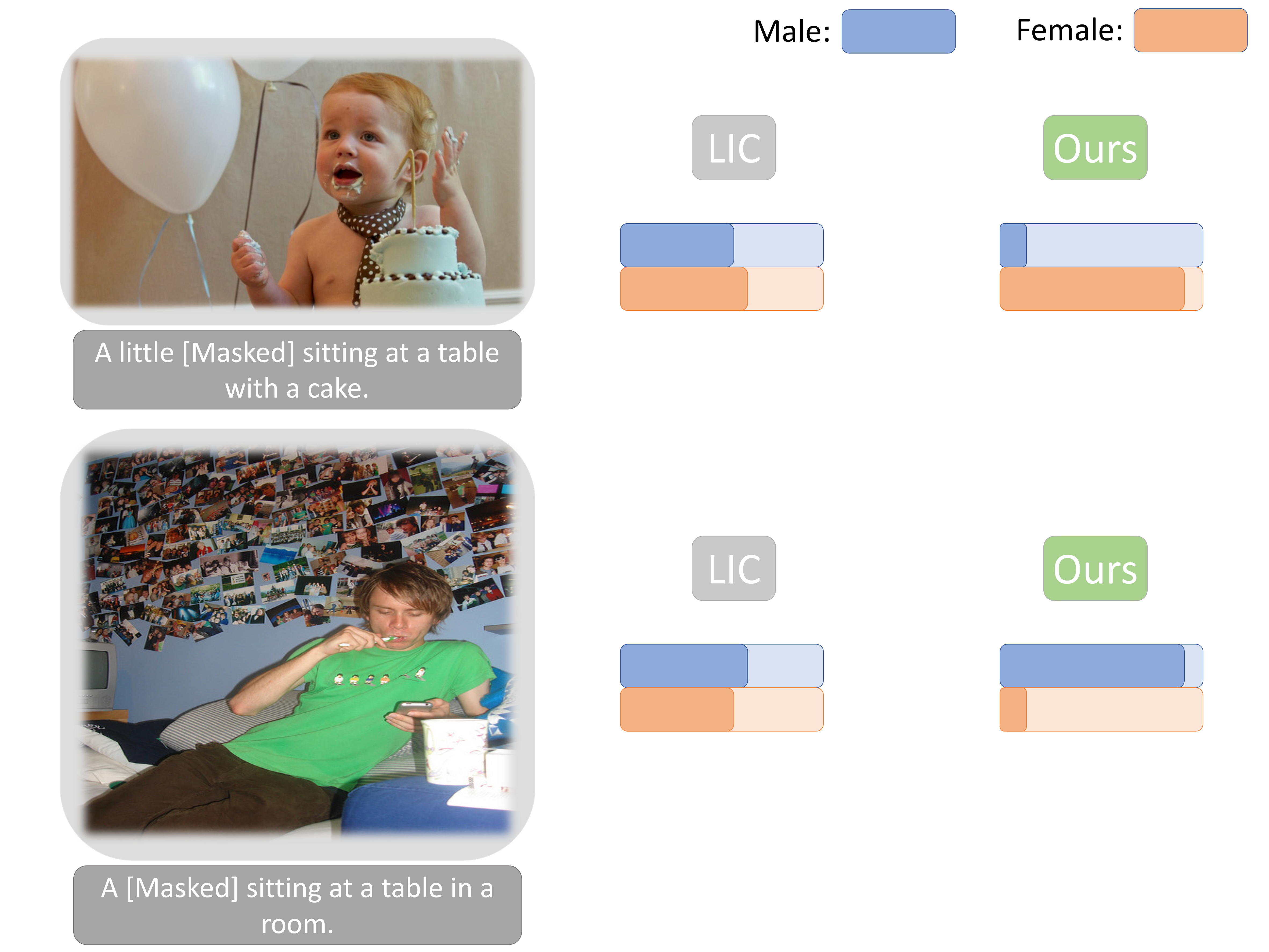}
\end{center}
\vspace{-0.5cm}
\caption{Qualitative samples that depict the captions generated by a biased model, accompanied by LIC and our scores.}
\label{fig_qualitative_samples}
\vspace{-0.3cm}
\end{figure}

Figure \ref{fig_qualitative_samples} presents the captions generated by a biased model, accompanied by LIC and our scores. 
Notably, LIC scores remain nearly identical, indicating its inability to detect bias. 
In contrast, our metric effectively identifies a significant bias.
This observation suggests that our metric considers the indoor scenario and photos on the wall as it can access visual cues.

\subsection{Emotion Bias Benchmark on ArtEmis}

We benchmark our metric, {\papernameAbbrev}, in terms of the emotion bias across Artemis-V1 \cite{achlioptas2021artemis} and Artemis-V2 \cite{mohamed2022okay} on three variant of SAT, based on the emotion source, i.e., Vanilla-SAT, predicted EG-SAT, and voting EG-SAT.
The predicted EG-SAT utilizes a pre-trained model to predict the emotion from an image, while the voting EG-SAT uses a simple voting mechanism on the GT data.
We follow the same training configurations mentioned in \cite{achlioptas2021artemis}.

As shown in Table \ref{tab_emotion_benchmark}, based on our metric, the three captioning models amplify the bias, which is consistent with LIC \cite{hirota2022quantifying}.
Surprisingly, Artemis-V2 suffers from severe emotion bias despite its success in generating more distinctive and specific captions and solving the unbalancing problem in the emotion distribution in Artemis-V1.
Our bias scores are higher than LIC scores, as we use a different scoring function, where the random guess score, which indicates an unbiased model, is 50\% and 25\%, respectively.
\subsection{Racial Bias Benchmark on COCO}

\begin{table}
    \centering
    \captionsetup{font=scriptsize}
    \caption{The bias amplification results for the racial attribute on MS-COCO dataset.}
    \scalebox{0.8}{
    \begin{tabular}{c c c c c}
    \toprule[1.5pt]
        \textbf{Method} & \multicolumn{2}{c}{\textbf{LIC} $\downarrow$} & \multicolumn{2}{c}{\textbf{Ours} $\downarrow$} \\ 
		 & \textbf{LSTM} & \textbf{BERT} & \textbf{SAT} & \textbf{GRIT}  \\ 
        \cmidrule(r){1-1}
        \cmidrule(r){2-3}
        \cmidrule(r){4-5}
        NIC \cite{vinyals2015show}& 5.7 & -0.29 & 5.19 & 2.97 \\
        SAT \cite{xu2015show} & 4.5 & 3.32 & 5.86 & 5.71 \\
        FC \cite{rennie2017self} & 7.6 & 3.61 & 5.53 & 4.43 \\
        Att2in \cite{rennie2017self} & 8.6 & 4.69 & 5.69 & 5.21 \\
        UpDn \cite{anderson2018bottom} & 7.8 & 7.06 & 5.33 & 4.05 \\
        Trans \cite{vaswani2017attention} & 6.1 & 3.86 & 5.21 & 3.49 \\
        OSCAR \cite{li2020oscar} & 5.9 & 5.49 & 5.63 & 4.3 \\
        NIC+ \cite{hendricks2018women} & 7.6 & 7.1 & 5.93 & 6.34 \\
        NIC+Eq \cite{hendricks2018women} & 7.2 & 8.28 & 5.4 & 4.17 \\
        \bottomrule[1.5pt]
    \end{tabular}
    }
    \label{tab_racial_coco}
\end{table}
We benchmark our metric, {\papernameAbbrev}, in terms of the racial bias across a wide range of different captioning models, against the current SOTA bias metric LIC \cite{hirota2022quantifying}.
We follow the same training configurations mentioned in \cite{hirota2022quantifying} for a fair comparison.
We measure the racial metric on the COCO dataset using two different image captioners, i.e., SAT \cite{xu2015show}, and GRIT \cite{nguyen2022grit}.
Based on our metric, all captioning models amplify the bias, where the bias score is positive.
As shown in Table \ref{tab_racial_coco}, the best model, i.e., the lowest bias score, based on our metric is NIC.
In contrast, the worst model in terms of amplifying racial bias is NIC+.

%% file: sections/6_conclusion.tex
\section{Conclusion}

In this paper, we revisit the existing fairness metrics in image captioning and dissect their limitations. Accordingly, we find existing learnable metrics suffer from inconsistency against classifiers variations besides ignorance of the multi-modality nature of image captioning.
This motivates us to introduce a novel metric termed {\papernameAbbrev}, which considers the visiolinguistic nature of the image captioning task while estimating the bias.
Thanks to the proposed prompt-based image captioner, we reformulate the bias metric as a text generation objective that better matches the underlying pre-training objective of the image captioner we assess instead of using a limited language classifier.
We apply our metric across 11 different image captioning techniques on three different datasets, i.e., MS-COCO caption dataset, Artemis V1, and Artemis V2, and on three different protected attributes, i.e., gender, race, and emotions.
Finally, we propose {\benchnameAbbrev} to compare the human alignment with our metric against others, where, our metric shows significant superiority over the recent bias metric; LIC, 80\% and 54\%, respectively.